\documentclass[journal, 10pt]{IEEEtran}

\ifCLASSINFOpdf
\else
\fi

\usepackage[utf8]{inputenc}
\usepackage{graphicx}
\usepackage{comment}
\usepackage{amsmath,amssymb} %
\usepackage{color}

\usepackage{booktabs}
\usepackage{multirow}
\usepackage{bm}
\usepackage[font=normalsize,caption=false]{subfig}
\usepackage{caption}
\graphicspath{{./images/}}
\newcommand{\Tref}[1]{Table~\ref{#1}}
\newcommand{\Eref}[1]{Eq.~(\ref{#1})}
\newcommand{\Fref}[1]{Fig.~\ref{#1}}
\newcommand{\Sref}[1]{Sec.~\ref{#1}}
 
\newcommand{\etal}[1]{\textit{et al.}}
\newcommand{\red}[1]{\textcolor{red}{\textbf{#1}}}
\newcommand{\blue}[1]{\textcolor{blue}{#1}}

\usepackage{grffile}  %
\usepackage{times}
\usepackage{url}
\usepackage{hyperref}

\usepackage{xcolor}
\usepackage{soul}

\begin{document}
\title{Learning from Synthetic Shadows \\ for Shadow Detection and Removal}

\author{
    Naoto~Inoue,~\IEEEmembership{Student Member,~IEEE,} and~Toshihiko~Yamasaki,~\IEEEmembership{Member,~IEEE}%
    \thanks{
    The work of N. Inoue is financially supported by The University of Tokyo - NEC AI scholarship. T. Yamasaki is partially financially supported by JSPS KAKENHI Grant Number JP19K22863.}
    \thanks{The authors are with the Department of Information and Communication Engineering, The University of Tokyo, Bunkyo, Tokyo 113-86, Japan. (email: yamasaki@hal.t.u-tokyo.ac.jp)}
    \thanks{Copyright \copyright 2021 IEEE. Personal use of this material is permitted. Permission from IEEE must be obtained for all other uses, in any current or future media, including reprinting/republishing this material for advertising or promotional purposes, creating new collective works, for resale or redistribution to servers or lists, or reuse of any copyrighted component of this work in other works.}
}

\markboth{IEEE TRANSACTIONS ON CIRCUITS AND SYSTEMS FOR VIDEO TECHNOLOGY, VOL. *, NO. **, **, 2021}
{INOUE \MakeLowercase{\textit{et al.}}: Learning from Synthetic Shadows}

\maketitle

\begin{abstract}
  Shadow removal is an essential task in computer vision and computer graphics.
  Recent shadow removal approaches all train convolutional neural networks (CNN) on real paired shadow/shadow-free or shadow/shadow-free/mask image datasets.
  However, obtaining a large-scale, diverse, and accurate dataset has been a big challenge, and it limits the performance of the learned models on shadow images with unseen shapes/intensities.
  To overcome this challenge, we present SynShadow, a novel large-scale synthetic shadow/shadow-free/matte image triplets dataset and a pipeline to synthesize it.
  We extend a physically-grounded shadow illumination model and synthesize a shadow image given an arbitrary combination of a shadow-free image, a matte image, and shadow attenuation parameters.
  Owing to the diversity, quantity, and quality of SynShadow, we demonstrate that shadow removal models trained on SynShadow perform well in removing shadows with diverse shapes and intensities on some challenging benchmarks.
  Furthermore, we show that merely fine-tuning from a SynShadow-pre-trained model improves existing shadow detection and removal models.
  Codes are publicly available at \url{https://github.com/naoto0804/SynShadow}.
\end{abstract}


\section{Introduction}
Shadows are prevalent in nature.
Detecting and manipulating shadows is essential in many computer vision and computer graphics downstream tasks~\cite{xu2005insignificant,shih2009exemplar,khan2015automatic}.
Research on shadow removal has advanced drastically by learning-based methods~\cite{guo2012paired,gryka2015learning,vicente2017leave}, especially by convolutional neural networks (CNN)~\cite{khan2015automatic,qu2017deshadownet,wang2018stacked,hu2019direction}.
These methods are usually trained on pairs of shadow and shadow-free images of identical scenes in a supervised manner.
In order to make these pairs, we must first take a photo of a scene with shadows and then take another photo after removing the occluder.
Additionally, we should manually or automatically annotate a binary mask indicating the presence of shadows if the model requires mask information.
This process dramatically limits the number and variety of collectible scenes and can cause noisy/biased supervision.
Shadow removal models learned from such data do not generalize well to broad real-world shadow images, as discussed in~\cite{hu2019mask}.
Besides, there is no guarantee that shadow-free regions are unchanged under the sunlight conditions, making the paired training images unreliable.

Some approaches have been proposed to overcome this challenge.
The first approach~\cite{hu2019mask} learns the model on unpaired shadow and shadow-free images.
It is prevalent in many image-to-image translation tasks (e.g., CycleGAN~\cite{zhu2017unpaired}), but the optimization is under-constrained, and shadow models learned from unpaired data often perform poorly.
The second approach~\cite{le2019shadow} augments the existing paired triplets datasets, but the variety of images produced by this approach is limited because it just slightly darken or enlighten shadows given a triplet.
The third approach~\cite{xiaodong2020towards} obtains a new shadow image given a combination of shadow-free and mask image pairs.
However, the variety of generated shadows is limited, especially in terms of shadow intensity. This is because the mapping between them is learned fully in a data-driven way from the existing shadow datasets, which is a severe bottleneck, as we have mentioned above.

\begin{figure}[t]
  \centering
  \includegraphics[width=\hsize]{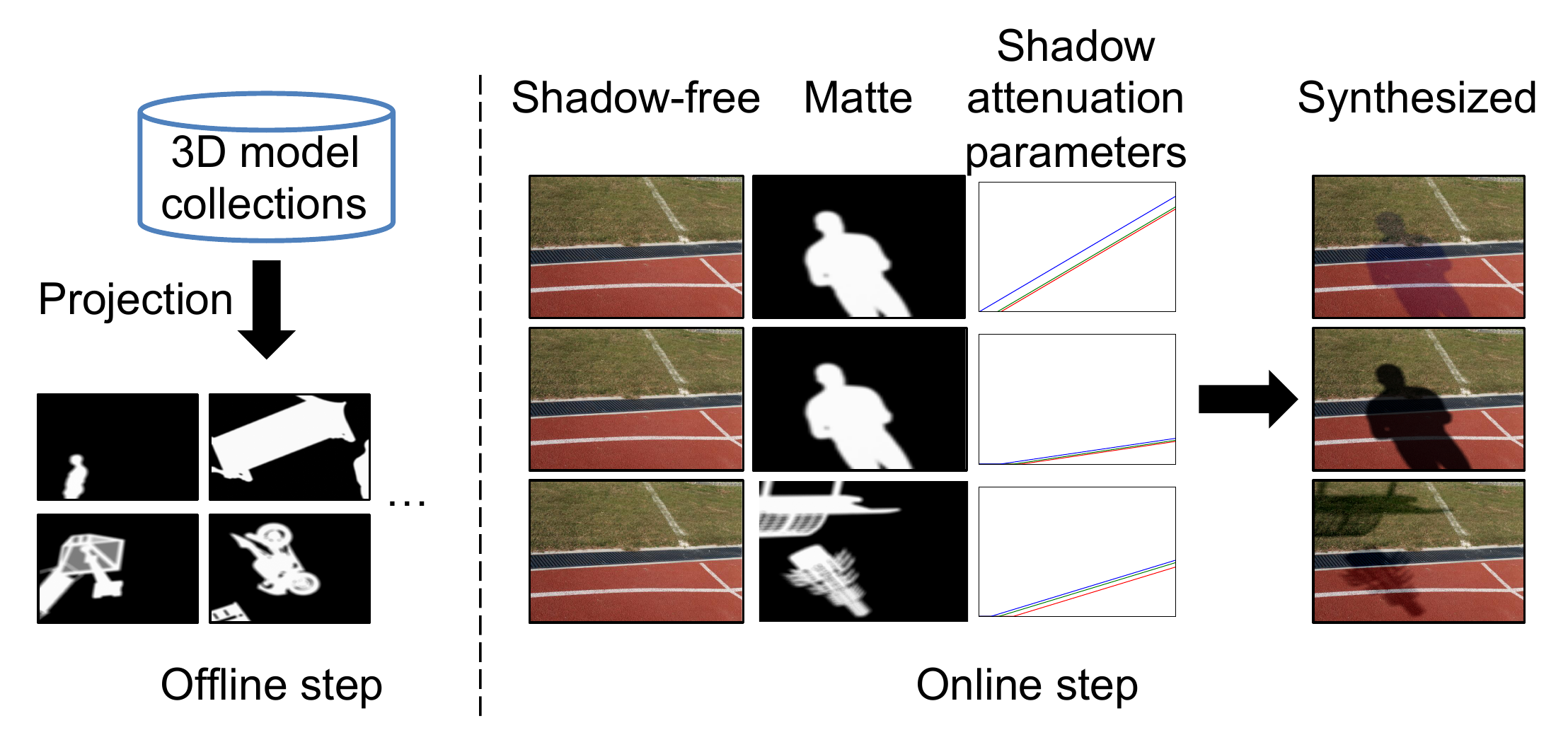} %
  \caption{
      Overview of our shadow synthesis pipeline.
      It can efficiently synthesize diverse and realistic shadow/shadow-free/matte image triplets.
      The triplet can be obtained from an arbitrary combination of a background image, shadow shape, and shadow attenuation property.
      Note that matte is \textbf{not} binary.
  }
  \label{fig:overview}
\end{figure}

In this paper, we tackle the mentioned challenge in shadow removal by generating a large-scale, diverse, yet realistic shadow/shadow-free/matte triplets dataset for supervised learning.
As shown in \Fref{fig:overview}, in our pipeline, a shadow image is composed of three components; (i) a shadow-free image as a background, (ii) a grayscale matte image indicating where the shadows are observed, and (iii) parameters indicating the shadow attenuation property.
We find that regarding these components to be independent of each other is essential compared to the previous composition approach~\cite{xiaodong2020towards}.
For (iii), we extend a physically-grounded shadow illumination model~\cite{shor2008shadow}.
We transform the model so that simply randomly sampling a set of parameters of it enables us to obtain diverse but realistic shadow attenuation.

Similar to the condition of existing shadow removal datasets, we assume all the occluder objects are outside the camera view in practice.
We separate the generation process into two independent parts, shadow matte generation (offline) and shadow composition using a shadow illumination model extending~\cite{shor2008shadow} (online), as shown in \Fref{fig:overview}.
In the offline step, we render shadow matte images from publicly available 3D models.
In the online step, we randomly sample the three components and then synthesize a shadow image from them.
The online step is computationally very cheap, and we can obtain new and diverse triplets on-the-fly during the training of neural network models.

Based on the pipeline, we propose a new dataset called SynShadow, containing shadow/shadow-free/matte image triplets synthesized from rendered 10,000 matte images and about 1,800 background images.
We train a variety of shadow detection and removal models on it to demonstrate its usefulness.
We demonstrate that simply fine-tuning from a SynShadow-pre-trained model improves existing shadow detection and removal models.
We demonstrate robust shadow removal results on challenging USR~\cite{hu2019mask} dataset, outperforming both supervised and unsupervised learning models trained on existing datasets in a user study.
We provide empirical and detailed analysis of the synthesis pipeline's design and demonstrate its superiority over existing shadow synthesis methods.

In summary, our contributions are as follows:
\begin{itemize}
    \item We present SynShadow, a large-scale dataset of shadow/shadow-free/matte image triplets, and the pipeline to synthesize the diverse and realistic triplets.
    \item We demonstrate the usefulness of SynShadow in improving various existing CNN models for shadow detection/removal by fine-tuning and achieving robust shadow removal on more challenging inputs.
\end{itemize}

\section{Related Work}

\subsection{Shadow Removal Methods and Datasets}
\label{subsec:shadow_removal_methods}

Shadow removal is essential mainly in two aspects;
(i) it is useful as a pre-processing step for downstream applications that are not robust to shadow (e.g., document recognition~\cite{bako2016removing,kligler2018document,das2019dewarpnet,lin2020bedsr}) and
(ii) it will give a perceptually better photo to users (e.g., in portraits~\cite{zhang2020portrait}).
Classical approaches remove shadows via user interaction~\cite{gong2014interactive} or hand-crafted features~\cite{guo2012paired,finlayson2005removal,liu2011cast,yang2012shadow}.
The emergence of CNN enables shadow removal to learn from shadow/shadow-free image pairs, or shadow/shadow-free/mask triplets~\cite{qu2017deshadownet,wang2018stacked,hu2019direction} by regressing all the pixel values directly.
In this approach, researchers have been trying to efficiently fuse both global and local contexts such as a multi-stream network~\cite{qu2017deshadownet}, a two-stage approach~\cite{wang2018stacked}, a direction-aware attention mechanism~\cite{hu2019direction}, or a recurrent module~\cite{ding2019argan}.
In contrast to these works, SP+M~\cite{le2019shadow} proposes a shadow image decomposition-based approach.
First, given an input shadow image, SP+M predicts parameters of a physical shadow illumination model~\cite{shor2008shadow} to generate a relit image.
Second, it predicts a shadow matte layer to blend the input shadow image and the relit image to generate the final de-shadowed image.
We show that our SynShadow can improve the performance of both approaches.

The datasets for shadow removal that we have discussed about are constructed by taking a photo with shadows and then taking another photo of the scene by removing the occluder.
This approach suffers from several limitations, as discussed by~\cite{hu2019mask}.
First, it is a very time-consuming process, resulting in a limited number of unique scenes in the exiting shadow removal datasets.
Second, it limits the diversity of the collected shadow shapes since some occluder such as buildings and trees cannot be removed.
Third, the training pairs may have color/luminosity inconsistency or a slight shift in the camera view since the camera exposure, pose, and environmental lighting may change during the pair collection process.

Mask-ShadowGAN (MSGAN)~\cite{hu2019mask} proposes to learn shadow removal from unpaired and diverse shadow/shadow-free images by extending CycleGAN~\cite{zhu2017unpaired}.
CycleGAN was modified to learn a mapping from a set of a shadow-free image and a binary shadow mask to a single shadow image and vice versa.
However, learning from unpaired data is quite unstable, and even shadow-free regions are often retouched, resulting in unnatural/undesired color/texture shifts.
In contrast, we convert diverse shadow-free images into the paired triplets for supervised learning to cope with this limitation.

\subsection{Shadow Synthesis}
\label{subsec:related_work_shadow_synthesis}
Existing approaches for synthesizing diverse and realistic shadow images are divided into three.

The first approach is to render shadow/shadow-free image pairs directly using a 3D renderer.
Sidorov~\etal~~\cite{sidorov2019conditional} renders these pairs in urban landscape from a computer game.
However, these images are significantly different from popular benchmarks for shadow removal, SRD~\cite{qu2017deshadownet} and ISTD~\cite{wang2018stacked}, because most of the occluder in these images is inside the camera view.
Gryka~\etal~'s work~\cite{gryka2015learning} for user-guided shadow removal first loads it to a 3D renderer, places automatically-segmented silhouettes of real objects, and renders shadows on the background by ray-tracing.
However, their rendering results may be physically incorrect because the renderer has no access to the background image's material information.
We composite shadows on background images with the help of shadow illumination model~\cite{shor2008shadow} that directly models the relationship between shadow and shadow-free images.

The second approach is to augment existing real datasets, which we call \textit{shadow augmentation}.
Le~\etal~~\cite{le2019shadow} use the same shadow illumination model with our approach.
Specifically, given a shadow/shadow-free/binary mask triplet, they estimate the parameters of the shadow illumination model and only slightly perturb the shadow attenuation slope (by a scaling factor ranging from 0.8 to 1.2) to create slightly different shadow images.
However, there are two drawbacks.
First, the variety of the generated shadow images is not diverse since they only augment existing datasets and cannot generate the images from a new scene.
Second, they regard these parameters as being dependent of physical conditions in the scene of the given triplet.
The parameters obtained from one triplet cannot be transferred to another triplet.
On the other hand, using our proposed parameter randomization of the shadow illumination model, we can synthesize shadow/shadow-free/matte image triplets given arbitrary combinations of shadow-free background images and shapes of occluder objects.

The third approach is composing shadows given an arbitrary shadow-free image as a background, which we call \textit{shadow composition}.
Cun~\etal~~\cite{xiaodong2020towards} propose Shadow Matting GAN (SMGAN), which composes realistic shadows given a shadow-free image and a randomly sampled shadow mask using CNN.
However, there are two drawbacks.
First, SMGAN itself is learned from existing real shadow/shadow-free/mask image triplets, which also limits the variety of shadows that SMGAN can generate in terms of shadow intensity.
Second, SMGAN assumes a one-to-one correspondence between shadow-free/mask pair collections and shadow image collections for learning the mapping between them.
This is unrealistic because an unlimited number of shadow images with varying shadow intensity can be possible given a shadow-free/mask image pair.
In contrast, we generate hundreds of different shadow images of a scene with varying shadow intensity given the same pair.

There are some recent papers on shadow generation~\cite{liu2020arshadowgan,inoue2019learning,wang2020people}.
However, they are not applicable to our setting.
ARShadowGAN~\cite{liu2020arshadowgan} and RGB2AO~\cite{inoue2019learning} are for casting shadows caused by the occluder inside the camera view.
Wang~\etal~'s work~\cite{wang2020people} is limited to a scene where a video from a single viewpoint is available.

\subsection{Shadow Detection}
Traditional approaches for shadow detection develop a physical model based on illumination invariant assumption~\cite{finlayson2005removal,salvador2004cast,finlayson2009entropy,russell2017feature} or employ various hand crafted features~\cite{xu2005insignificant,guo2012paired,vicente2017leave,lalonde2010detecting,zhu2010learning,huang2011characterizes,panagopoulos2012simultaneous}.
Recent approaches use CNN~\cite{khan2015automatic,shen2015shadow,vicente2016large} and try to effectively capture global and local context~\cite{wang2018stacked,hu2019direction,nguyen2017shadow,zhu2018bidirectional}.
DSDNet~\cite{zheng2019distraction} proposes to mine hard-positive/negative regions in the existing dataset and develop losses and modules to attend these regions explicitly.
Being orthogonal to DSDNet, we synthesize a large-scale dataset including various hard-positive/negative regions and let the detection models learn from them implicitly.

Collecting datasets for shadow detection is essential since prevalent approaches are data-hungry.
Existing datasets are classified into three: (i) when the occluder is outside the camera view~\cite{qu2017deshadownet,wang2018stacked}, (ii) when the occluder is inside the camera view
\cite{vicente2016large,wang2020instance}, and (iii) mixture of both~\cite{hu2019revisiting}.
We show that our synthetic dataset is useful to improve the performance of shadow detection models for the case (i).

\section{SynShadow Pipeline}
\label{sec:synshadow_pipeline}
Our goal is to synthesize a large number of triplets $(\bm{x}^{s}, \bm{x}^{ns}, and \bm{m})$. $\bm{x}^{s} \in \mathbb{R}^{H \times W \times 3}$, $\bm{x}^{ns} \in \mathbb{R}^{H \times W \times 3}$, and $\bm{m} \in \mathbb{R}^{H \times W}$ indicate a shadow image, a shadow-free image, and a shadow matte, respectively.
$H$ and $W$ indicate the height and width of the images, respectively.
All the values in $\bm{x}^{s}$, $\bm{x}^{ns}$, $\bm{m}$ are between zero and one.
Our pipeline consists of two steps, as shown in \Fref{fig:overview}: shadow matte generation and shadow composition.
We assume that we can compose any shadow matte on random background images only when occluders are outside the image.

In \Sref{subsec:shadow_illumination_model}, we briefly summarize the shadow illumination model used in~\cite{le2019shadow,shor2008shadow} formulating the relation between shadow and shadow-free pixels.
In \Sref{subsec:shadow_synthesis}, we explain how to synthesize shadow given an arbitrary combination of $(\bm{x}^{ns}$, $\bm{m})$ using the illumination model.
In \Sref{subsec:matte_synthesis}, we explain how to obtain diverse shadow matte images.

\subsection{Preliminary: Shadow Illumination Model}
\label{subsec:shadow_illumination_model}
In~\cite{shor2008shadow}, the model is derived from the image formation equation proposed in~\cite{barrow1978recovering}:
\begin{equation}
    I(p, \lambda) = L(p, \lambda)R(p, \lambda),
\end{equation}
where $I(p, \lambda)$ is a scalar value indicating the intensity of the light reflected from the point $p$ at the wavelength $\lambda$.
$L$ and $R$ indicate the luminance and the reflectance, respectively.
The model assumes a single primary light source and an ambient light as the sources of illumination.
It is further assumed that the shadow is cast solely by the primary light source (e.g., the sun in outdoor scenes).
If $p$ is shadow-free (i.e., lit), $L$ at $p$ can be expressed as a sum of two terms:
\begin{equation}
    L(p, \lambda) = L^{d}(p, \lambda) + L^{a}(p, \lambda),
\end{equation}
where $L^{d}$ is the direct illumination and $L^{a}$ is the ambient illumination.
Therefore, the intensity $I^{lit}(p, \lambda)$ we see at $p$ is expressed as
\begin{equation}
    I^{lit}(p, \lambda) = L^{d}(p, \lambda)R(p, \lambda) + L^{a}(p, \lambda)R(p, \lambda). \label{eq:i_lit}
\end{equation}
In the case where some objects occlude the primary light source from the point $p$ and shadows are casted, the reflected intensity $I^{dark}$ is
\begin{equation}
    I^{dark}(p, \lambda) = a(p)L^{a}(p, \lambda)R(p, \lambda), \label{eq:i_shadow}
\end{equation}
where $a(p)$ is a factor indicating the attenuation of the ambient illumination by the occluder.
$a(p)$ is assumed to have roughly the same spectral distribution from all incident directions. It is thus assumed to be independent of $\lambda$.
By combining \Eref{eq:i_lit} and \Eref{eq:i_shadow}, the relation between $I^{lit}$ and $I^{dark}$ is formulated as follows:
\begin{equation}
    I^{lit}(p, \lambda) = L^{d}(p, \lambda)R(p, \lambda) + \frac{1}{a(p)}I^{dark}(p, \lambda). \label{eq:shadow_to_lit}
\end{equation}
When a photo is taken, the actual color at a pixel in the photo corresponding to the 3D point $p$ in the scene is obtained by integrating the both sides of \Eref{eq:shadow_to_lit} with the camera's spectral response functions.
This operation is assumed not to change the affine nature of the relationship between the shadowed and illuminated intensities. Thus, the relation between $I^{lit}$ and $I^{dark}$ at any pixel for the $k$-th color channel ($k$ = 0, 1, and 2 for red, green, and blue) are expressed as follows:
\begin{equation}
    I^{lit}_{k} = \alpha_{k} + \gamma~I^{dark}_{k}, \label{eq:shor_affine_model}
\end{equation}
where both $\alpha_{k}$ and $\gamma$ are scalar values.
Thus, the shadow attenuation property of a 3D scene is represented by $\alpha_{k}$ and $\gamma$. These parameters depend on camera and scene properties, such as the material of surfaces and lighting conditions.
Shor~\etal~~\cite{shor2008shadow} demonstrated that it applies well to actual photos, even though it is not perfect due to the presence of noise and the variations in the reflectance of surfaces.

\subsection{Shadow Synthesis}
\label{subsec:shadow_synthesis}
Here we describe the proposed shadow synthesis pipeline.
Let a shadow free image be $\bm{x}^{ns}$.
We first obtain an image $\bm{x}^{dark} \in \mathbb{R}^{H \times W \times 3}$, where all the pixels are sheltered and have the same attenuation property.
We use the affine model in \Eref{eq:shor_affine_model} to compute $\bm{x}^{dark}$ from $\bm{x}^{ns}$:
\begin{align}
    & x^{ns}_{ijk} = \alpha_{k} + \gamma~x^{dark}_{ijk} \\
    \iff & x^{dark}_{ijk} = \frac{1}{\gamma}~x^{ns}_{ijk} - \frac{\alpha_{k}}{\gamma}, \label{eq:shor_affine_model_uniform}
\end{align}
where $i$ and $j$ indicate indices for row and column axes, respectively, for all the images.

The final image with shadows in some regions, $\bm{x}^{s}$, is obtained by composing $\bm{x}^{ns}$ and $\bm{x}^{dark}$ by alpha composition using the shadow matte $\bm{m}$ as the alpha factor;
\begin{align}
    x^{s}_{ijk} &= (1 - m_{ij}) x^{ns}_{ijk} + m_{ij} x^{dark}_{ijk}. \label{eq:alpha_composition}
\end{align}

\begin{figure}[t]
  \centering
  \includegraphics[width=0.4\hsize]{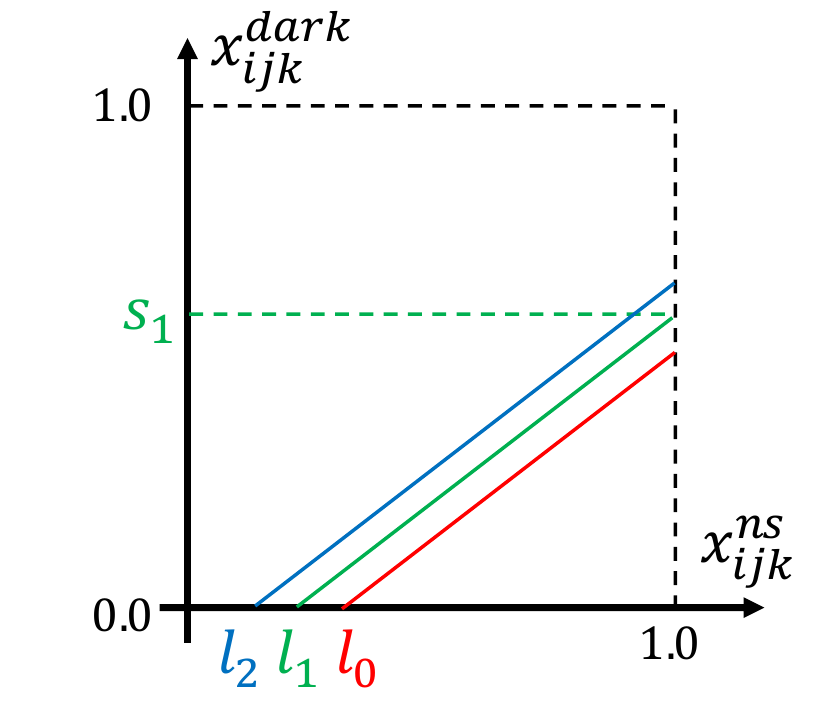}
  \caption{
      Parameters $(l_{0}, l_{1}, l_{2}, s_{1})$ that we introduce for analysis.
  }
  \label{fig:params_space}
\end{figure}

\begin{figure}[t]
  \centering
  \includegraphics[width=\hsize]{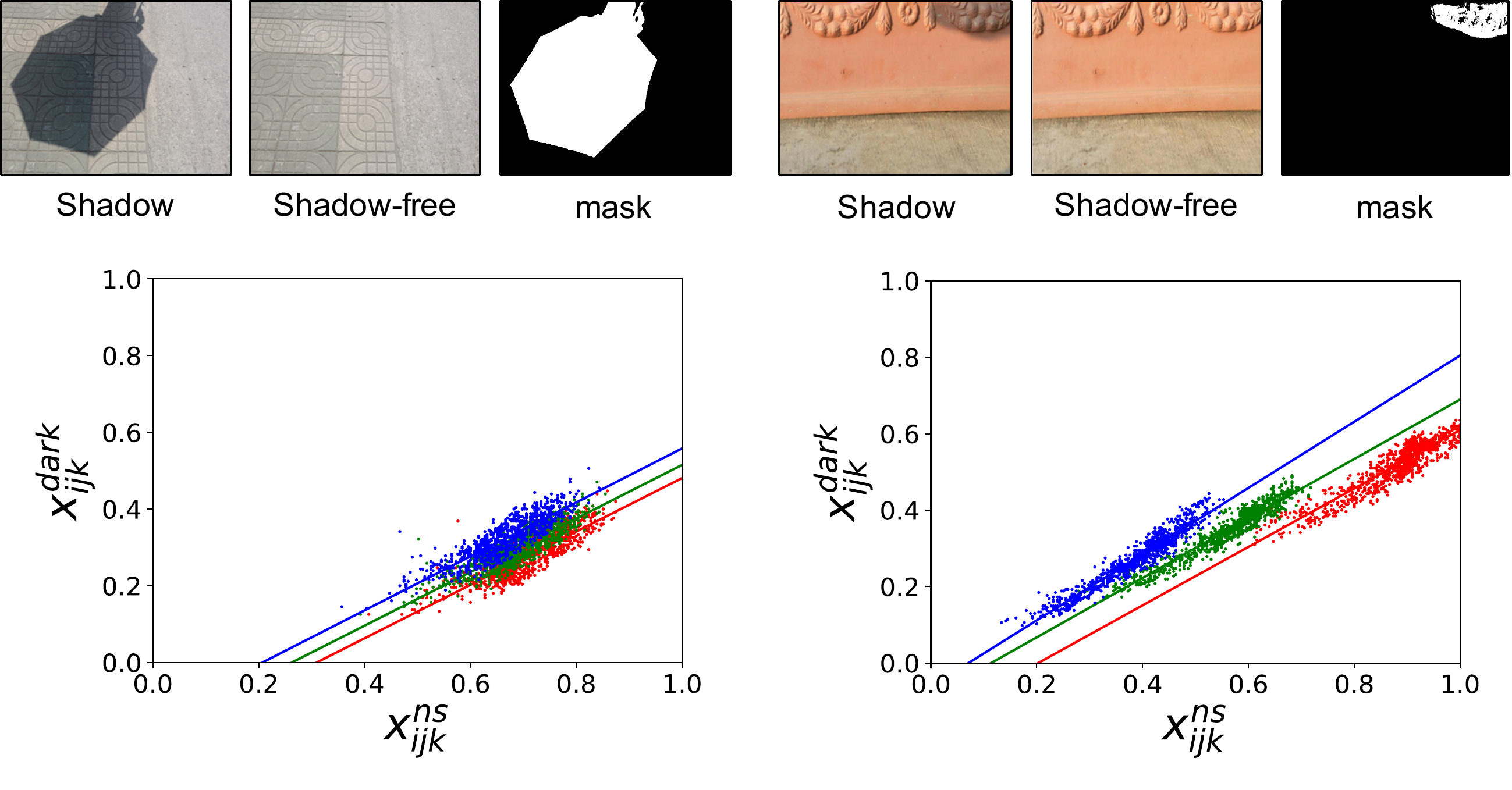}
  \caption{
      Visualization of the shadow attenuation for each RGB channel in the ISTD+ and SRD+ training set.
      The left and right sides are the visualization of an example triplet from ISTD+ and SRD+ training set, respectively.
      We fit linear functions regressing $x^{dark}_{ijk}$ from $x^{ns}_{ijk}$ for each channel and show the estimated functions as the lines.
  }
  \label{fig:attenuation_visualization}
\end{figure}

We next discuss how to sample $\alpha_{k}$ and $\gamma$ to produce both plausible and diverse shadows.
This is unknown and non-trivial.
We provide observation for understanding the relation of $\alpha_{k}$ and $\gamma$.
To clearly explain it, we convert $\alpha_{k}$ and $\gamma$ to four parameters $(l_{0}, l_{1}, l_{2}, s_{1})$ as shown in \Fref{fig:params_space}.
Formally, this is written as: $s_{1} = \frac{1 - \alpha_{1}}{\gamma}, l_{k} = \alpha_{k}$.
Given a shadow/shadow-free/mask image triplet in the ISTD~\cite{le2019shadow} and SRD~\cite{qu2017deshadownet} datasets, we visualize the relationship between the intensity of pixels in the shadowed region in \Fref{fig:attenuation_visualization}.
We confirm that the shadow illumination model in~\cite{shor2008shadow} holds.
Besides, we observe that $(l_{0}, l_{1}, l_{2})$ correlate to each other, and often the relation is $l_{0} > l_{1} > l_{2}$.
We conjecture that this is because the ambient light is from the blueish sky in outdoor scenes, which is also suggested in~\cite{huang2011characterizes}.

Based on this observation, we parametrize the shadow illumination model that covers a plausible but diverse range of shadows.
Inspired by domain randomization~\cite{tobin2017domain}, we obtain a set of parameters, where each of them is easy to sample from an interpretable yet straightforward prior distribution and independent of each other.
Since $l_{k}$'s are dependent of each other based on the observation, we further introduce $\Delta l_{0} = l_{0} - l_{1}$ and $\Delta l_{2} = l_{2} - l_{1}$ and sample $(l_{1}, s_{1}, \Delta l_{0}, \Delta l_{2})$.
We employ a uniform distribution $\mathcal{U}(a, b)$ for both $l_{1}$ and $s_{1}$.
$(a,b)=(0.0, 0.25)$ and $(a,b)=(0.1, 0.9)$ are employed for $l_{1}$ and $s_{1}$, respectively.
We employ normal distribution $\mathcal{N}(\mu, \sigma)$ for both $\Delta l_{0}$ and $\Delta l_{2}$.
$(\mu, \sigma) = (0.05, 0.025)$ and $(\mu, \sigma) = (-0.05, 0.025)$ are employed for $\Delta l_{0}$ and $\Delta l_{2}$ for R and B channel, respectively.

Finally, $\bm{x}^{dark}$ is computed as follows:
\begin{equation}
    x^{dark}_{ijk} = \begin{cases}
        \frac{s_{1}}{1.0 - l_{1}}(x^{ns}_{ijk} - l_{k}) & \text{if}~~x^{ns}_{ijk} - l_{k} \geq 0.0, \\
        0 & \text{if}~~x^{ns}_{ijk} - l_{k} < 0.0.\\
    \end{cases}
\end{equation}
If the slope of the shadow attenuation $\frac{s_{1}}{1.0 - l_{1}}$ is larger than one, it is unnatural.
In that case we re-sample $(l_{1}, s_{1}, \Delta l_{0}, \Delta l_{2})$ to stabilize the quality of the sampled shadow images.

\begin{figure}[t]
  \centering
  \includegraphics[width=\hsize]{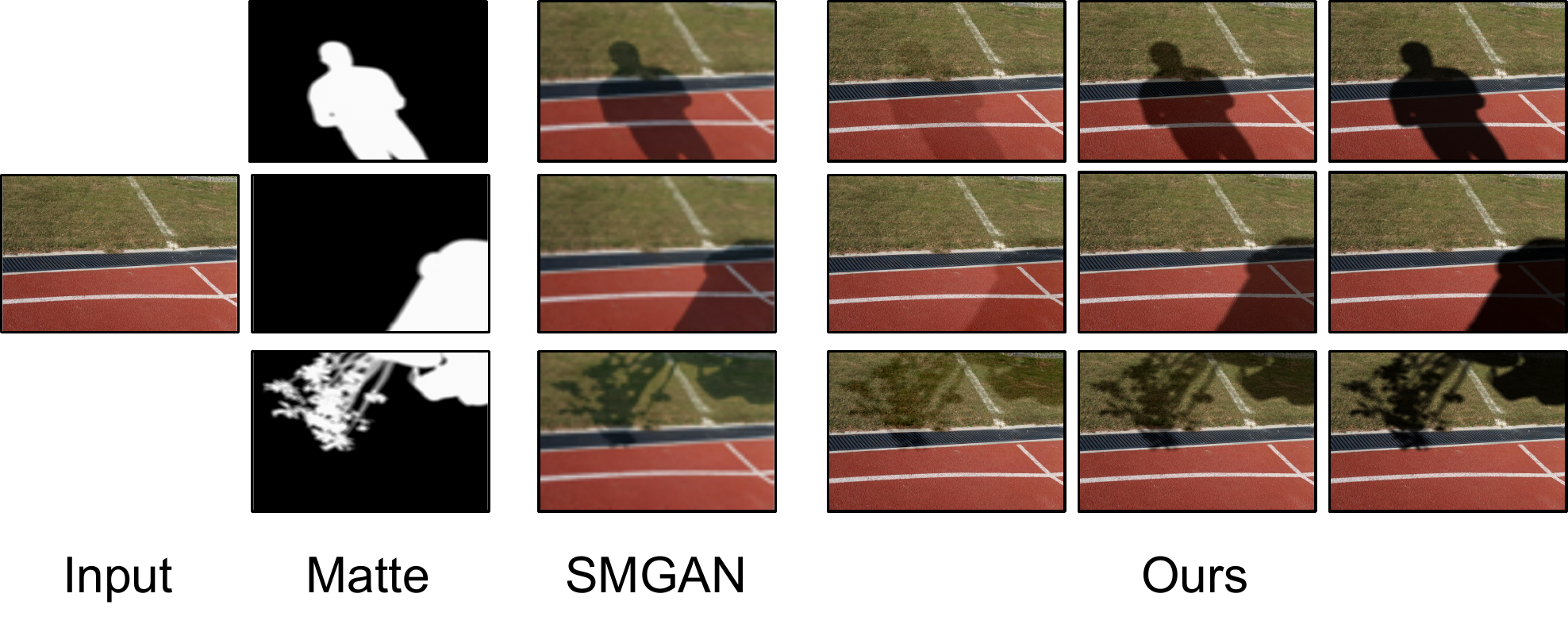}
  \caption{
      Difference between Shadow Matting GAN (SMGAN)~\cite{xiaodong2020towards} and our shadow composition model.
  }
  \label{fig:diff_smgan_affine}
\end{figure}

{
\newcommand{\fig}[1]{\frame{\includegraphics[width=0.18\hsize]{images/dataset_comp/#1}}}
\begin{figure}[t]
    \centering
    \begin{tabular}{@{}ccc|c@{}}
        \fig{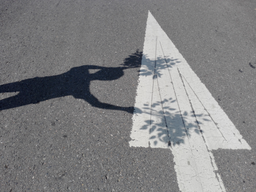} & \fig{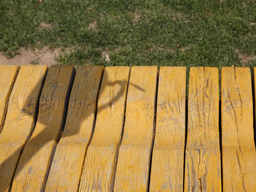} & \fig{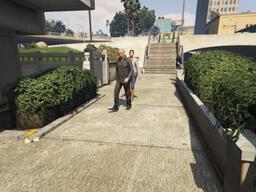} & \fig{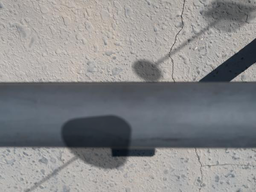} \\
        \fig{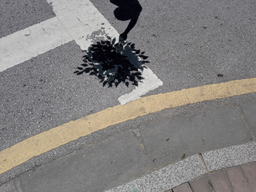} & \fig{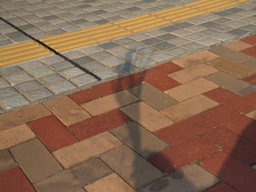} & \fig{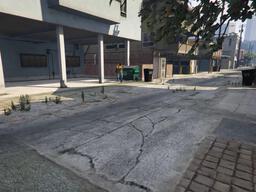} & \fig{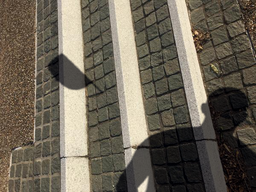} \\
        \fig{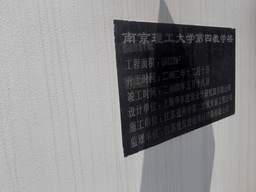} & \fig{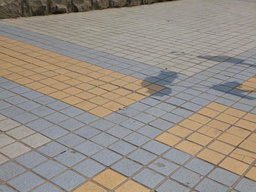} & \fig{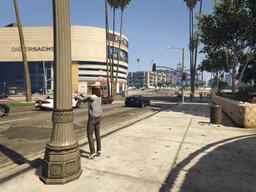} & \fig{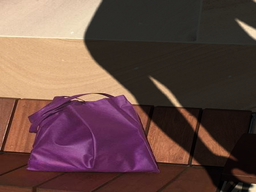} \\
        ISTD~\cite{wang2018stacked} & SRD~\cite{qu2017deshadownet} & GTAV~\cite{sidorov2019conditional} & SynShadow \\
    \end{tabular}
    \caption{
        Comparison of the datasets for shadow removal.
        Shadows in GTAV are mostly caused by occluder objects inside the camera, while shadows in ISTD, SRD, and SynShadow are caused by those outside the camera.
    }
    \label{fig:dataset_comp}
\end{figure}
}

We show the result of our shadow composition in \Fref{fig:diff_smgan_affine}.
While SMGAN only generates a single shadow image given an input image and a matte, our composition model is able to create various shadow images.
We show the comparison of the datasets for shadow removal in \Fref{fig:dataset_comp}.
Although the composited shadow may not fully match the background geometry, we can obtain realistic and diverse shadows.

\subsection{Shadow Matte Generation}
\label{subsec:matte_synthesis}

\begin{figure}[t]
  \centering
  \includegraphics[width=0.5\hsize]{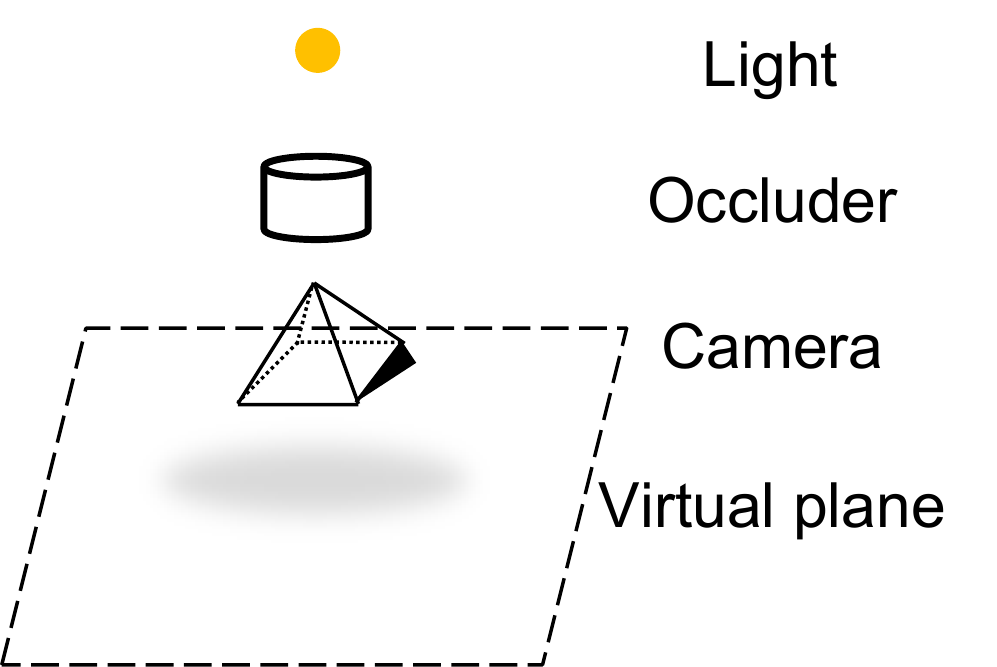}
  \caption{Overview of the occluder projection.}
  \label{fig:occluder_projection}
\end{figure}

{
\newcommand{\fig}[1]{\frame{\includegraphics[width=0.18\hsize]{images/srd_overlap/#1}}}
\begin{figure}[t]
    \centering
    \begin{tabular}{@{}cccccc@{}}
        \fig{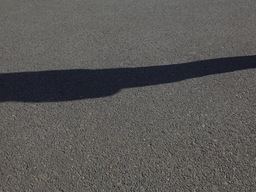} & \fig{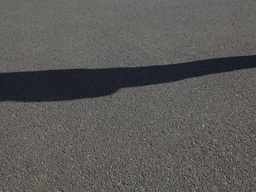} & \fig{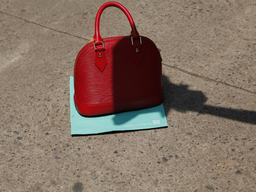} & \fig{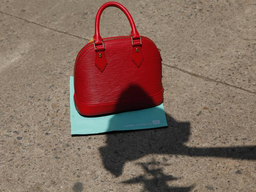} \\
    \end{tabular}
    \caption{
        Examples of the scene overlap between the train-test split in SRD~\cite{qu2017deshadownet}.
        The images in the odd and even columns are from the training and testing set, respectively.
        The left pair seems to be near duplicates.
        The right pair share exactly the same background.
    }
    \label{fig:srd_overlap}
\end{figure}
}

We obtain a shadow matte $\bm{m}$, where $m_{ij} = 1$ if the pixel is inside the umbra, $0 \leq m_{ij} \leq 1$ if the pixel is in the penumbra, and $m_{ij} = 0$ otherwise.
We put a light, occluder objects, a camera, and a virtual plane in Blender\footnote{https://www.blender.org/} as shown in \Fref{fig:occluder_projection}.
We capture the virtual plane through the camera, obtain the amount of light reaching each point, and normalize the amount to obtain the shadow matte.
Note that all the occluder objects are assumed to be outside the camera view, and thus they are not apparent in the final shadow/shadow-free/matte images.
Therefore, we need the geometric relation of the light, the occluder, the camera, and the virtual plane.
We randomly scale, translate, and rotate each component to generate a diverse shadow matte.
We sample occluder and light as follows.

\textbf{Occluder}: %
We use AMASS~\cite{mahmood2019amass} and ShapeNet~\cite{shapenet2015}, which are collections of publicly available 3D models as the occluder objects.
We ignore the material or other information of the models and only use their geometric information.
AMASS contains a large number of captured human 3D mesh sequences.
We randomly sample 3D meshes from AMASS to obtain 3D human meshes.
For common objects, we randomly sample 3D meshes from 26 object categories that are often seen in outdoor scenes from ShapeNet.
In a single capture, we sample up to two models, where each model is either a human or an ordinary object.

\textbf{Light}:
By changing the radius of the spherical light in Blender, a different penumbra can be obtained in the same geometrical configuration.
We randomly sample various radii to obtain a diverse variety of penumbras for the same scene.
Note that there is no need to randomize the brightness of the light.
We will randomize this parameter separately later using the shadow illumination model.

\section{Experiments on Shadow Removal}
\label{sec:exp_shadow_removal}
\subsection{Datasets, Models, and Evaluation Metrics}
We evaluate the potential of SynShadow comparing with other datasets by using several state-of-the-art algorithms.
Note that we are not proposing novel techniques for shadow removal but a new dataset.

\subsubsection{Datasets}
We employed three datasets, ISTD+, SRD+, and USR, for evaluating shadow removal performance.

\textbf{ISTD+}:
ISTD+~\cite{le2019shadow} consists of shadow/shadow-free/mask image triplets.
Note that the mask is represented in a binary format.
It has 1,330 and 540 triplets for training and testing, respectively.
It is a color-adjusted version of ISTD~\cite{wang2018stacked} to cope with a color inconsistency issue between the shadow images and their corresponding shadow-free images caused by the triplet collection process.
Although all the prior works used the original ISTD, \cite{le2019shadow} showed that ISTD is not appropriate for evaluating shadow removal methods fairly.

\textbf{SRD+}:
SRD+ has 2,675 and 406 shadow/shadow-free/mask image triplets for training and testing, respectively.
SRD+ is based on SRD~\cite{qu2017deshadownet}.
We found that the original train-test split of SRD is inappropriate since images coming from the identical background are both in the training and testing sets, as shown in \Fref{fig:srd_overlap}.
Therefore, we re-split SRD so that there is no overlap of scenes in the two sets and removed near-duplicate images.
SRD initially consists of shadow/shadow-free image pairs.
Mask for shadow region is extracted following \cite{xiaodong2020towards}.

\textbf{USR}:
USR~\cite{hu2019mask} consists of unpaired 2,445 shadow images and 1,770 shadow-free images.
It is more challenging than ISTD+/SRD+ because it contains diverse shadows.
It consists of thousands of different diverse scenes, while SRD and ISTD are only comprised of hundreds of scenes.
The shadow images are split into 1,956 and 489 images for training and testing, respectively.

We also describe the detailed configuration of SynShadow.

\textbf{SynShadow}:
We generate 10,000 shadow matte images to generate SynShadow as described in \Sref{sec:synshadow_pipeline}.
The shadow-free images are obtained from USR.

\subsubsection{Models}
We employed three approaches.

\textbf{Supervised learning}:
Supervised learning models are trained on paired shadow/shadow-free images such as ISTD+, SRD+, and SynShadow.
We mainly employed the best recent models, decomposition-based SP+M~\cite{le2019shadow} and regression-based DHAN~\cite{xiaodong2020towards}.
We also employed regression-based DSC.
For a fair comparison, when we train models on SynShadow, training details such as learning schedule and hyper-parameters are similar to those used for training on ISTD+/SRD+.
Note that the codes of ST-CGAN~\cite{qu2017deshadownet}, DeshadowNet~\cite{wang2018stacked}, and ARGAN~\cite{ding2019argan} are not publicly available, and we could not evaluate their performance on ISTD+/SRD+ unfortunately.

\textbf{Unsupervised learning}:
Unsupervised learning models are trained on unpaired shadow/shadow-free images.
Following \cite{hu2019mask}, we tested MSGAN~\cite{hu2019mask} and CycleGAN~\cite{zhu2017unpaired}.
We also trained these models on SRD+/ISTD+ in an unpaired manner for a more fair comparison with supervised learning models.

\textbf{Traditional methods}:
We tried Guo~\etal~~\cite{guo2012paired} and Gong~\etal~~\cite{guo2012paired}.
Note that Gong~\etal~~\cite{gong2014interactive} is an interactive method that requires the user's manual input.

\subsubsection{Evaluation Metrics}
We followed all the prior works on shadow removal in quantitative evaluation.
We used root-mean-square error (RMSE) in LAB color space between the ground truth and predicted shadow-free images.
RMSE is reported for all pixels (ALL).
Additionally, RMSE is reported for only shadow pixels (S) and only non-shadow pixels (NS).
Smaller RMSE indicates better performance.

\subsection{Experiments on ISTD+/SRD+ datasets}
\label{subsec:exp_rem_istd+_srd+}
We demonstrate how we can use SynShadow to improve the supervised shadow removal models on the existing shadow removal datasets, ISTD+ and SRD+.
We considered two scenarios for using SynShadow;

\textbf{Zero-shot}:
We trained the models on SynShadow and evaluated them on ISTD+/SRD+ test set.

\textbf{Fine-tuning}:
We pre-trained the models on SynShadow, fine-tuned them on ISTD+/SRD+ train set, and evaluated them on ISTD+/SRD+ test set.

\begin{table}[t]
  \normalsize
  \setlength{\tabcolsep}{2pt}
  \caption{
      RMSE comparison with the state-of-the-art methods in the LAB color space.
      * indicates an interactive method that requires a user's manual input as additional supervision during testing.
      I/S is short for ISTD+ \textit{or} SRD+, so that the dataset for training and evaluation is similar.
      SS is short for SynShadow.
  }
  \label{tbl:comparison_removal_seen}
  \centering
  \begin{tabular}{@{}cccccccc@{}} \toprule
      \multicolumn{2}{c}{Tested on} & \multicolumn{3}{c}{ISTD+} & \multicolumn{3}{c}{SRD+} \\
      \multicolumn{2}{c}{Metrics} & S & NS & ALL & S & NS & ALL \\ \midrule
      Methods & Trained on & & & & & & \\ \midrule
      \multicolumn{8}{l}{(a)~\textit{Traditional}} \\
      Guo~\etal~~\cite{guo2012paired} & - & 22.3 & 4.3 & 7.1 & 24.3 & 6.5 & 10.3 \\
      Gong~\etal~~\cite{gong2014interactive}* & - & 14.4 & \textbf{3.4} & 5.1 & 18.5 & 3.9 & 7.0 \\ \midrule
      \multicolumn{8}{l}{(b)~\textit{Unsupervised}} \\
      MSGAN~\cite{hu2019mask} & USR & 24.7 & 6.9 & 9.9 & 30.3 & 11.0 & 15.1 \\
      CycleGAN~\cite{zhu2017unpaired} & USR & 25.6 & 7.4 & 10.2 & 29.9 & 11.5 & 15.4 \\
      MSGAN~\cite{hu2019mask} & I/S & 14.1 & 7.6 & 8.6 & 16.5 & 7.6 & 9.5 \\
      CycleGAN~\cite{zhu2017unpaired} & I/S & 14.3 & 7.9 & 8.9 & 17.5 & 7.5 & 9.6 \\ \midrule
      \multicolumn{8}{l}{(c)~\textit{Supervised}} \\
      SP+M~\cite{le2019shadow} & SS & 11.3 & 3.6 & 4.9 & 11.6 & 4.1 & 5.7 \\
      DHAN~\cite{xiaodong2020towards} & SS & 9.7 & 4.0 & 4.9 & 13.8 & 5.1 & 6.9 \\
      SP+M~\cite{le2019shadow} & I/S & 8.5 & 3.6 & 4.4 & 12.2 & \textbf{3.4} & 5.3 \\
      DHAN~\cite{xiaodong2020towards} & I/S & 7.4 & 4.8 & 5.2 & 12.7 & 5.9 & 7.4 \\
      DSC~\cite{hu2019direction} & I/S & 8.3 & 4.6 & 5.2 & 16.2 & 5.8 & 8.0 \\ \midrule
      \multicolumn{8}{l}{(d)~\textit{Supervised, pre-trained on SynShadow}} \\
      SP+M~\cite{le2019shadow} & I/S & \textbf{6.9} & \textbf{3.4} & \textbf{4.0} & \textbf{10.9} & 3.6 & \textbf{5.2} \\

      DHAN~\cite{xiaodong2020towards} & I/S & 6.6 & 4.2 & 4.6 & 10.6 & 5.6 & 6.6 \\ \bottomrule
  \end{tabular}
\end{table}

\subsubsection{Quantitative Evaluation}
The results are shown in \Tref{tbl:comparison_removal_seen}.
We discuss the results in each scenario.

\textbf{Zero-shot}:
Models trained on SynShadow (the first two rows in group c) outperformed unsupervised learning (group b) and traditional methods (group a) that do not use the ISTD+/SRD+ dataset for training.
Furthermore, the model trained on SynShadow is almost compatible with most of the supervised learning models (the last three rows in group c) trained and evaluated on a similar dataset.

\textbf{Fine-tuning}:
When we fine-tuned SynShadow-pre-trained models (group d) for each specific dataset, we constantly obtained the best results.

\subsubsection{Qualitative Evaluation}
{
\newcommand{\figa}[1]{\frame{\includegraphics[width=0.11\hsize]{images/comparison_removal/test_istd+/#1}}}
\newcommand{\figb}[1]{\frame{\includegraphics[width=0.11\hsize]{images/comparison_removal/test_srd+/#1}}}

\begin{figure*}[t]
    \centering
    \setlength{\tabcolsep}{3pt}
    \begin{tabular}{ccccccc}
        \figa{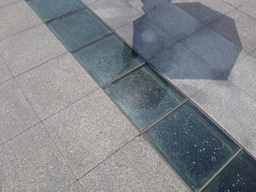} & \figa{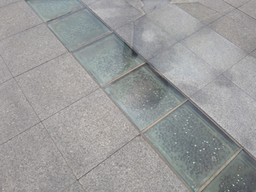} & \figa{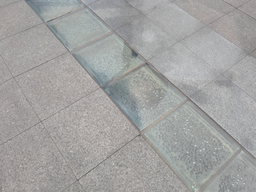} & \figa{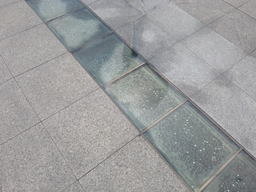} & \figa{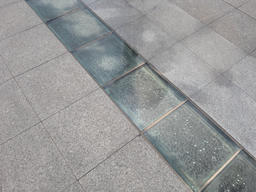} & \figa{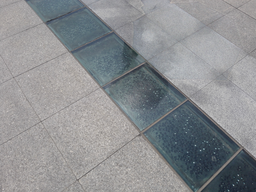} & \figa{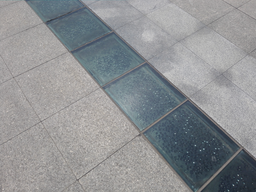} \\
        \figa{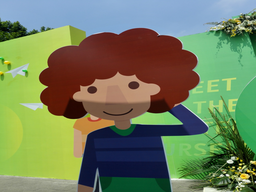} & \figa{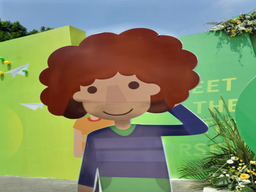} & \figa{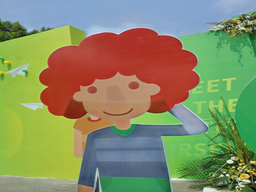} & \figa{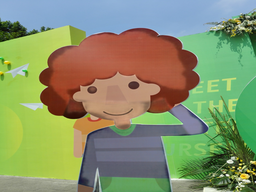} & \figa{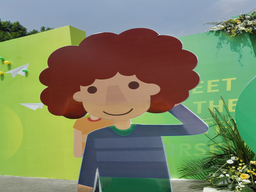} & \figa{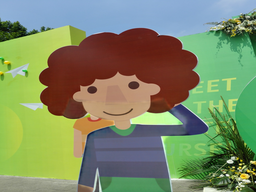} & \figa{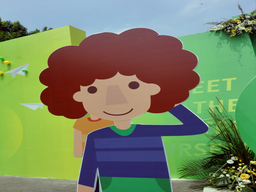} \\ \midrule
        \figb{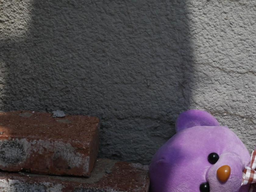} & \figb{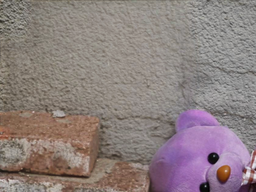} & \figb{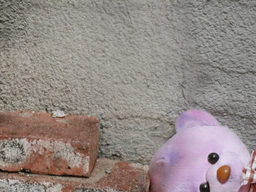} & \figb{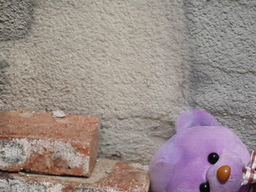} & \figb{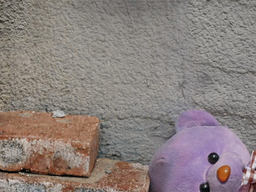} & \figb{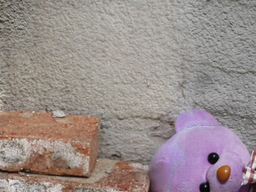} & \figb{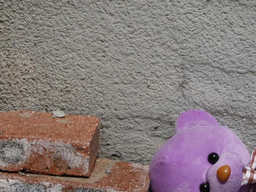} \\
        \figb{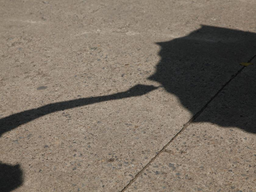} & \figb{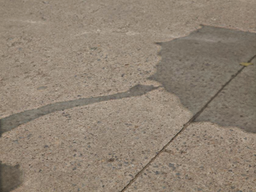} & \figb{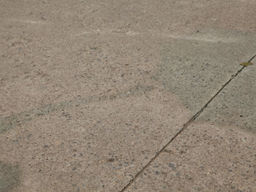} & \figb{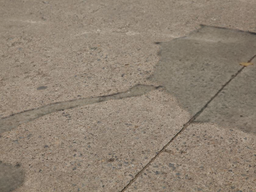} & \figb{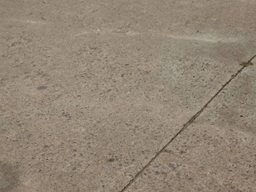} & \figb{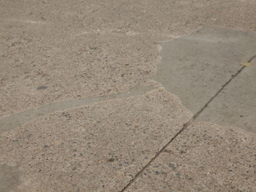} & \figb{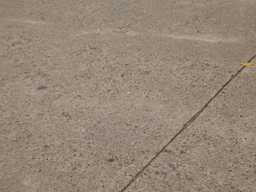} \\
        Input & DSC~\cite{hu2019direction} & DHAN~\cite{xiaodong2020towards} & SP+M~\cite{le2019shadow} & DHAN-ft & SP+M-ft & Ground truth \\
    \end{tabular}
    \caption{
        Qualitative comparison of shadow removal models.
        Results in the top and bottom two rows are from models trained and evaluated on ISTD+ and SRD+, respectively.
        DHAN-ft and SP+M-ft indicate DHAN and SP+M pre-trained on SynShadow and later fine-tuned on each dataset.
    }
    \label{fig:comparison_removal_istd}
\end{figure*}
}

In \Fref{fig:comparison_removal_istd}, we show the visual comparison of some supervised learning models discussed in \Tref{tbl:comparison_removal_seen}.
The fine-tuned model's improvement is attributed to not modifying non-shadow areas (the 1st and 2nd rows) and improved relit estimation (the 3rd and 4th rows).

\subsubsection{Transferability of Shadow Removal Datasets}
\begin{table*}[t]
  \normalsize
  \setlength{\tabcolsep}{2pt}
  \caption{
      Comparison by changing the training/fine-tuning dataset in shadow removal.
      Top two results in each setting are highlighted in \red{red} and \blue{blue}, respectively.
  }
  \label{tbl:ablation_finetuning_strategy}
  \centering
  \begin{tabular}{@{}cccccccccccccc@{}} \toprule
      Tested on & \multicolumn{6}{c}{ISTD+} & & \multicolumn{6}{c}{SRD+} \\
      Model & \multicolumn{3}{c}{SP+M} & \multicolumn{3}{c}{DHAN} & & \multicolumn{3}{c}{SP+M} & \multicolumn{3}{c}{DHAN} \\
      Metrics & S & NS & ALL & S & NS & ALL & & S & NS & ALL & S & NS & ALL \\ \midrule
      \multicolumn{7}{l}{\textit{Training}} \\
      GTAV & 26.9 & 4.5 & 8.0 & 28.1 & 10.9 & 13.6 & GTAV & 27.8 & 4.1 & 9.1 & 32.5 & 7.2 & 12.6 \\
      SRD+ & 14.9 & \blue{3.8} & 5.6 & 12.2 & 6.3 & 7.2 & SRD+ & \blue{12.2} & \red{3.4} & \red{5.3} & \red{12.7} & \blue{5.9} & \blue{7.4} \\
      SynShadow & \blue{11.3} & 3.6 & \blue{4.9} & \blue{9.7} & \red{4.0} & \red{4.9} & SynShadow & \red{11.6} & \blue{4.1} & \blue{5.7} & \blue{13.8} & \red{5.1} & \red{6.9} \\
      ISTD+ & \red{8.5} & \red{3.6} & \red{4.4} & \red{7.4} & \blue{4.8} & \blue{5.2} & ISTD+ & 14.9 & 4.5 & 6.7 & 16.6 & 8.1 & 9.9 \\ \midrule
      \multicolumn{7}{l}{\textit{Fine-tuning}} \\
      GTAV$\rightarrow$ISTD+ & 7.8 & 3.6 & 4.2 & \blue{7.2} & \red{4.2} & \blue{4.7} & GTAV$\rightarrow$SRD+ & \blue{11.9} & \blue{3.6} & \blue{5.4} & 13.0 & \blue{5.6} & 7.2 \\
      SRD+$\rightarrow$ISTD+ & \blue{7.6} & \blue{3.5} & \blue{4.1} & \blue{7.2} & \blue{4.7} & 5.1 & ISTD+$\rightarrow$SRD+ & \blue{11.9} & \red{3.4} & \red{5.2} & \blue{11.8} & \red{5.5} & \blue{6.8} \\
      SynShadow$\rightarrow$ISTD+ & \red{6.9} & \red{3.4} & \red{4.0} & \red{6.6} & \red{4.2} & \red{4.6} & SynShadow$\rightarrow$SRD+ & \red{10.9} & \blue{3.6} & \red{5.2} & \red{10.6} & \blue{5.6} & \red{6.6} \\ \bottomrule
  \end{tabular}
\end{table*}

We conducted more detailed analysis on transferability of many shadow removal datasets listed in \Fref{fig:dataset_comp}.
We again consider both zero-shot and fine-tuning setting in \Tref{tbl:ablation_finetuning_strategy}.

\textbf{Zero-shot}:
The result is shown in the upper half of \Tref{tbl:ablation_finetuning_strategy}.
When the domain of datasets for training and evaluation is different, the models trained on SynShadow perform best.
Surprisingly, the DHAN model trained on SynShadow even performs better than those trained on similar domain of datasets for training and evaluation.
We conjecture that regression-based models may overfit to a small dataset.

\textbf{Fine-tuning}:
The result is shown in the lower half of \Tref{tbl:ablation_finetuning_strategy}.
We can clearly see the advantage of pre-training on SynShadow, compared to the other datasets.

\subsubsection{Ablation Study on Parameters Randomization}
\begin{table}[t]
  \normalsize
  \setlength{\tabcolsep}{2pt}
  \caption{
      Ablation study on design of randomizing parameters in the shadow illumination models.
      SP+M is trained on each variant for quantitative evaluation.
      Top two results in each setting are highlighted in \red{red} and \blue{blue}, respectively.
  }
  \label{tbl:ablation_affine_model_variants}
  \centering
  \begin{tabular}{@{}ccccccc@{}} \toprule
      Tested on & \multicolumn{3}{c}{ISTD+} & \multicolumn{3}{c}{SRD+} \\
      Metrics & S & NS & ALL & S & NS & ALL \\ \midrule
      Color jitter & 39.3 & 4.6 & 10.1 & 44.0 & 4.6 & 13.0 \\
      Color jitter (dark) & \red{10.8} & 4.7 & 5.6 & \red{11.4} & 4.7 & \blue{6.1} \\
      Gamma correction & 23.3 & 3.9 & 6.9 & 33.6 & \blue{4.1} & 10.3 \\ \midrule
      Independent & 13.4 & 4.0 & \blue{5.4} & 12.6 & 4.2 & 6.0 \\
      Zero intercepts & 14.9 & 3.7 & \blue{5.4} & 13.2 & 4.2 & \blue{6.1} \\
      Similar intercepts & 23.3 & \red{3.5} & 6.7 & 18.9 & \red{4.0} & 7.1 \\
      Non-biased intercepts & 12.1 & 5.5 & 6.6 & 12.8 & 5.9 & 7.3 \\
      Proposed & \blue{11.3} & \blue{3.6} & \red{4.9} & \blue{11.6} & \blue{4.1} & \red{5.7} \\ \bottomrule
  \end{tabular}
\end{table}

We conduct ablation study on how to generate shadow images.

First, we tested various settings on how to generate the affine shadow illumination model parameters $(l_{0}, l_{1}, l_{2}, s_{1})$.
Five variants including proposed method are tested:

\begin{itemize}
    \item Independent params:
        $(l_{0}, l_{1}, l_{2}, s_{1})$, are randomized independently.
        Note that the range of each parameter is similar to our setting.

    \item Zero intercepts:
        $l_{1}$ is fixed at 0.0.
        Note that $l_{0}$ and $l_{2}$ are still randomized with respect to $l_{1}$.

    \item Similar intercepts:
        The relation between $(l_{0},l_{1},l_{2})$ is $l_{0}=l_{1}=l_{2}$. In this case, the shadow attenuation is assumed to be similar in RGB channels.

    \item Non-biased intercepts:
        We set $\mu=0.0$ in randomizing parameters.

    \item Proposed:
        This is based on our proposal, and we did not try any of the above modifications.
\end{itemize}

Second, we tested some possible approaches for generating shadow images directly by simple low-level image filtering operations.
In all the cases, we used the same shadow-free and matte images as those used for obtaining SynShadow.

\begin{itemize}
    \item Color jitter:
        We tried color jitter, which is often used for image augmentation, for each RGB channel to make the shadow images.
        In this case, attenuation in each channel is assumed to be independent of each other.
        We followed a paper for portrait shadow manipulation~\cite{zhang2020portrait}, where color jitter is formulated as a $3 \times 3$ color correction matrix.

    \item Color jitter (dark):
        We found that the original implementation of \cite{zhang2020portrait} sometimes makes the shadowed region brighter.
        We limited the range of color jitter so that the shaded region is always darker for a fair comparison.

    \item Gamma correction:
    We tried gamma correction to make the shadow images.
    We compute $x^{dark}_{ijk} = (x^{ns}_{ijk})^{y}$, where $1.5 \leq y \leq 3.0$ is randomly sampled to produce diverse shadows.
\end{itemize}

Comparison among all the variants is shown in \Tref{tbl:ablation_affine_model_variants}.
The proposed methods performed much better than the other variants in both datasets.
This result stresses the importance of carefully setting the range in our proposed randomization of parameters and selecting the proper shadow model.

\subsection{Experiments on USR dataset}
\label{subsec:exp_on_usr}

\subsubsection{User Study}
We performed a user study since there is no paired supervision for quantitative evaluation in the USR dataset.
The models trained on SynShadow are compared with the best approaches from supervised/unsupervised/traditional methods.
30 shadow images are randomly chosen from the testing set.
We displayed results from a single input on a webpage in random order at the same time.
Unlike the user study in the MSGAN paper~\cite{hu2019mask}, we showed the original input as a reference so that the evaluators can notice the difference between the results and the input.
We asked Amazon Mechanical Turk workers to rate each result on a scale from 1 (bad) to 5 (excellent).
300 votes are obtained from 10 users, and we report the mean and standard deviation of the ratings in \Tref{tbl:user_study_usr}.
Since comparing results from an arbitrary combination of methods and datasets at once is very difficult, we conducted several experiments.

\textbf{Fixed supervised learning model, different datasets}:
We fixed the model and changed the dataset for supervised learning to demonstrate the significance of SynShadow.
First, we show the result using the SP+M model in \Tref{tbl:user_study_usr_spm}.
The result of the model trained on SynShadow obtained much better ratings compared to those trained on the other datasets.
Second, we show the result using the DHAN model in \Tref{tbl:user_study_usr_dhan}.
For comparison with an existing shadow composition method, SMGAN~\cite{xiaodong2020towards}, we report the results using additional datasets.
Following \cite{xiaodong2020towards}, we combined images generated by SMGAN with the ISTD+ and SRD+ datasets, and obtained SM-ISTD+ and SM-SRD+, respectively.
Although SM-ISTD+, SM-SRD+, and SynShadow internally use USR shadow-free images, the model trained on SynShadow performed the best.

\textbf{Comparison with unsupervised learning or traditional approaches}:
SP+M and DHAN models trained on SynShadow are compared with traditional approaches and unsupervised learning approaches in \Tref{tbl:user_study_usr_unsup_trad}.
Both methods obtained much better ratings than the compared approaches.
\Fref{fig:comparison_removal_usr} shows the visual comparisons.
The previous best approach for the USR dataset, MSGAN~\cite{hu2019mask}, tends to fail to focus only on the shadow region, resulting in drastic unwanted changes in textures in the shadow-free areas.
Therefore, supervised learning on the diverse data, even though they are synthetic, is essential for robust shadow removal.

\textbf{Comparison with shadow augmentation}:
In \Tref{tbl:user_study_usr_diff_aug}, we compare our approach with the shadow augmentation approach~\cite{le2019shadow} using the SP+M model.
Specifically, we consider the following settings in addition to the original ISTD+ dataset:

\begin{itemize}
    \item
        Augmented ISTD+: Following \cite{le2019shadow}, we augmented existing shadow/shadow-free/mask triplets in the original ISTD+ dataset.
        Please refer to the third paragraph in \Sref{subsec:related_work_shadow_synthesis} for details.
    \item
        SynShadow (BG-ISTD+): For unbiased evaluation, we used shadow-free images in the original ISTD+ dataset as the background for obtaining SynShadow.

\end{itemize}
Although \cite{le2019shadow}'s augmentation contributed to the improved rating, the model based on SynShadow obtained a much better rating, even when the shadow-free images are from the ISTD+ dataset.
This result further highlights the importance of adequately randomizing the shadow illumination model as we propose, compared to \cite{le2019shadow}.

\begin{table}[t]
  \centering
  \normalsize
  \setlength{\tabcolsep}{2pt}
  \caption{
      Comparison of models trained on different datasets.
      User study results on USR testing set is reported.
  }
  \label{tbl:user_study_usr}
  \subfloat[Model: SP+M~\cite{le2019shadow} \label{tbl:user_study_usr_spm}]{
      \begin{tabular}{cc} \toprule
          Trained on & Rating \\ \midrule
          ISTD+ & 2.00 $\pm$ 1.02 \\
          SRD+ & 1.52 $\pm$ 0.83 \\
          SynShadow & \textbf{3.00} $\pm$ 1.40 \\ \bottomrule
      \end{tabular}
  }
  \hspace{0.02\textwidth}
  \subfloat[Model: DHAN~\cite{xiaodong2020towards} \label{tbl:user_study_usr_dhan}]{
      \begin{tabular}{cc} \toprule
          Trained on & Rating \\ \midrule
          ISTD+ & 2.52 $\pm$ 1.18 \\
          SM-ISTD+ & 1.68 $\pm$ 0.96 \\
          SRD+ & 2.67 $\pm$ 1.17 \\
          SM-SRD+ & 2.43 $\pm$ 1.16 \\
          SynShadow & \textbf{3.07} $\pm$ 1.19 \\ \bottomrule
      \end{tabular}
  }
\end{table}

\begin{table}[t]
  \centering
  \normalsize
  \setlength{\tabcolsep}{2pt}
  \caption{
      Comparison with unsupervised learning and traditional approaches.
      User study results on USR testing set is reported.
      * indicates an interactive method that requires a user's manual input as additional supervision during testing.
      \label{tbl:user_study_usr_unsup_trad}
  }
  \begin{tabular}{cc} \toprule
      Method & Rating \\ \midrule
      Guo~\etal~~\cite{guo2012paired} & 1.71 $\pm$ 1.21 \\
      Gong~\etal~~\cite{gong2014interactive}* & 2.61 $\pm$ 1.24 \\
      MSGAN~\cite{hu2019mask} & 2.37 $\pm$ 1.37 \\
      DHAN (SynShadow) & \textbf{3.25} $\pm$ 1.20 \\
      SP+M (SynShadow) & 3.00 $\pm$ 1.45 \\ \bottomrule
  \end{tabular}
\end{table}

\begin{table}[t]
  \centering
  \normalsize
  \setlength{\tabcolsep}{2pt}
  \caption{
      Comparison with shadow augmentation approach proposed in \cite{le2019shadow}.
      User study results on USR testing set is reported.
  }
  \label{tbl:user_study_usr_diff_aug}
  \begin{tabular}{cc} \toprule
      Trained on & Rating \\ \midrule
      ISTD+ & 1.94 $\pm$ 1.05 \\
      Augmented ISTD+ & 2.12 $\pm$ 1.05 \\
      SynShadow (BG-ISTD+) & \textbf{2.81} $\pm$ 1.29 \\ \bottomrule
  \end{tabular}
\end{table}

{
\newcommand{\fig}[1]{\frame{\includegraphics[width=0.11\hsize]{images/comparison_removal/test_usr/#1}}}
\begin{figure*}[t]
    \centering
    \setlength{\tabcolsep}{3pt}
    \begin{tabular}{cccccc}
        \fig{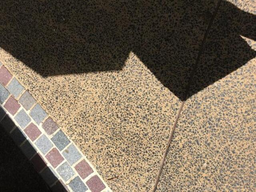} & \fig{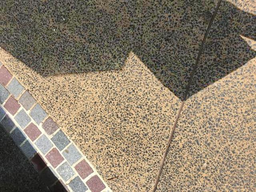} & \fig{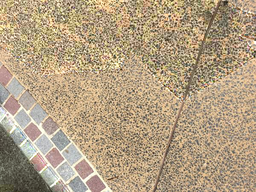} & \fig{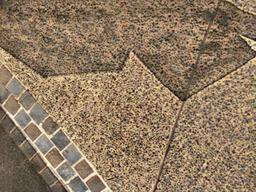} & \fig{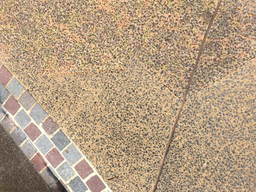} & \fig{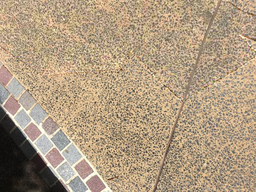}
         \\
        \fig{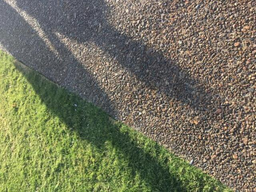} & \fig{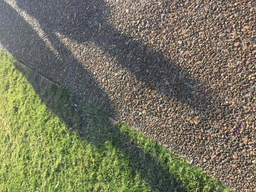} & \fig{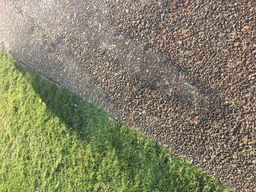} & \fig{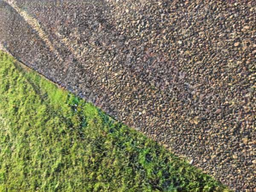} & \fig{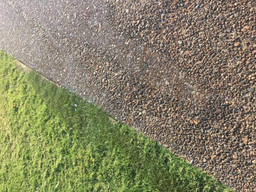} & \fig{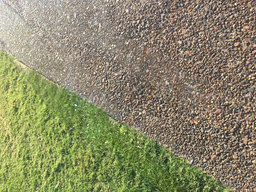}
         \\
        \fig{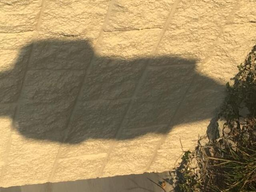} & \fig{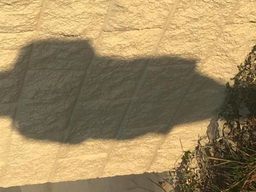} & \fig{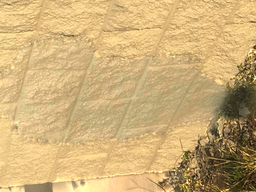} & \fig{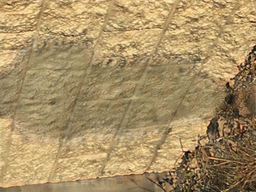} & \fig{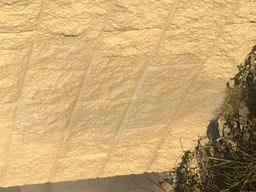} & \fig{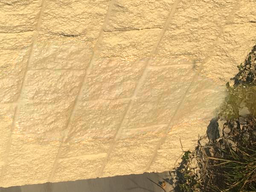}
         \\
        \fig{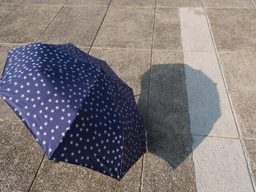} & \fig{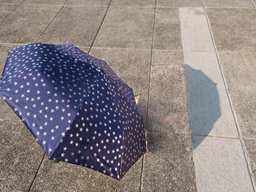} & \fig{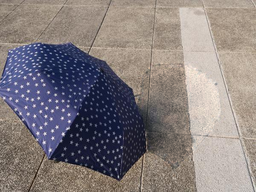} & \fig{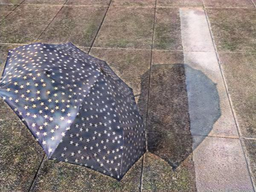} & \fig{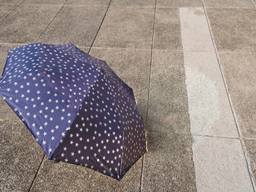} & \fig{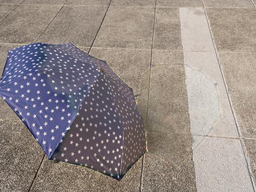}
         \\
         Input & Guo~\etal~~\cite{guo2012paired} & Gong~\etal~~\cite{guo2012paired} & MSGAN~\cite{hu2019mask} & DHAN (SS) & SP+M (SS) \\
    \end{tabular}
    \caption{
        Qualitative comparison of shadow removal methods on USR testing set.
        SS stands for SynShadow.
    }
    \label{fig:comparison_removal_usr}
\end{figure*}
}

{
\newcommand{\fig}[1]{\frame{\includegraphics[width=0.11\hsize]{images/ablation_mask_shadow/#1}}}
\begin{figure*}[t]
    \centering
    \setlength{\tabcolsep}{3pt}
    \begin{tabular}{cccccc}
        Shadow shape & ISTD+ & SRD+ & 3D models & 3D models & 3D models \\
        Composition & SM-ISTD+ & SM-SRD+ & SM-ISTD+ & SM-SRD+ & Our model \\
        \fig{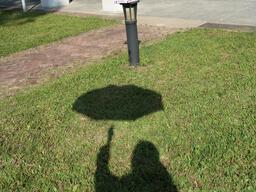} & \fig{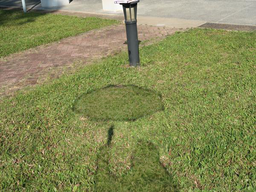} & \fig{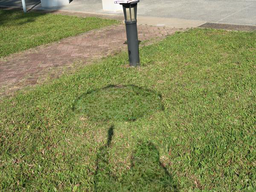} & \fig{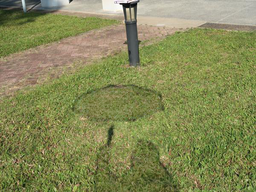} & \fig{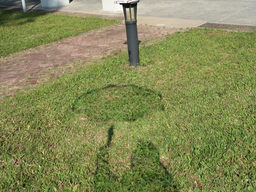} & \fig{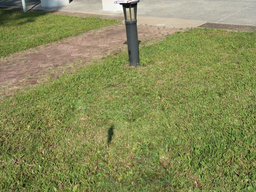} \\
        \fig{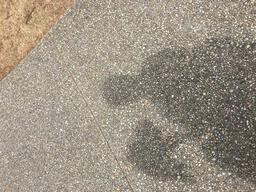} & \fig{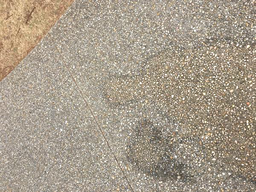} & \fig{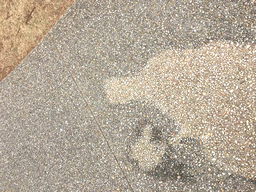} & \fig{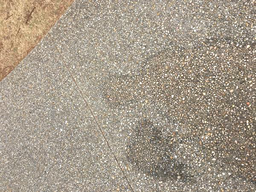} & \fig{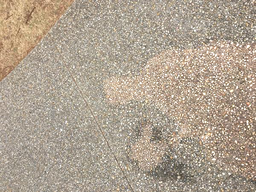} & \fig{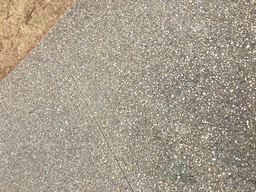} \\
        \fig{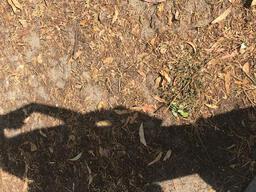} & \fig{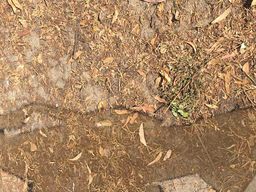} & \fig{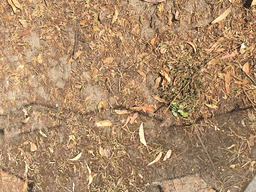} & \fig{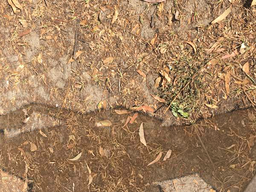} & \fig{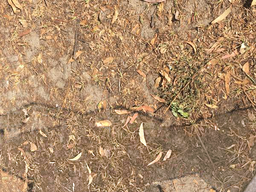} & \fig{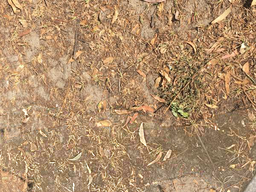} \\
        Input & \multicolumn{5}{c}{Removal results} \\
    \end{tabular}
    \caption{
        Ablation study on the choice of shadow composition model and source of shadow shape evaluated on USR testing set.
        As a shadow removal model, we used SP+M~\cite{le2019shadow}.
        SM-ISTD+ indicates SMGAN~\cite{xiaodong2020towards} trained on ISTD+ dataset.
    }
    \label{fig:ablation_composition_factors}
\end{figure*}
}

\subsubsection{Ablation Study on Components of Shadow Synthesis Framework}
For a more detailed analysis of how each randomized component of our shadow synthesis framework contributes to the improved removal result in the USR dataset, we compared the removal results by changing two factors of our pipeline:
\begin{itemize}
    \item Shadow shape: We tested shadow masks from the ISTD+ and SRD+ datasets, and shadow masks/matte from rendered 3D models.
    \item Composition: We tested the shadow illumination model and SMGAN~\cite{xiaodong2020towards} for comparison.
\end{itemize}
As shown in \Fref{fig:ablation_composition_factors}, choosing the shadow illumination model instead of SMGAN significantly boosts the shadow removal results. This result also suggests the importance of the proposed illumination randomization.

\section{Experiments on Shadow Detection}

\subsection{Evaluation Metrics}
Following the prior works, we used the balance error rate (BER)~\cite{vicente2017leave} for quantitative evaluation.
\begin{equation}
    BER = 1 - 0.5 \times (\frac{N_{tp}}{N_{p}} + \frac{N_{tn}}{N_{n}}),
\end{equation}
where $N_{tp}$, $N_{tn}$, $N_{p}$, and $N_{n}$ indicate the numbers of true positives, true negatives, shadow pixels, and non-shadow pixels, respectively.
A lower score indicates better performance.
BER is reported for all pixels (ALL).
Additionally, BER is reported for only shadow pixels (S) and only non-shadow pixels (NS).

\subsection{Models and Datasets}
For models, we used two models, DSDNet++ and BDRAR~\cite{zhu2018bidirectional}.
We modified DSDNet~\cite{zheng2019distraction}, and only use weighted binary cross-entropy loss for training, and did not use Distraction-aware Shadow (DS) loss proposed in the original DSDNet.
This is because we observed that it harms BER in our fine-tuning setting.
We call this variant DSDNet++.

For datasets, we use the ISTD testing set for the evaluation.
For the training/fine-tuning, we additionally consider the SBU dataset~\cite{vicente2016large} consisting of 4085 and 638 images for training and evaluation, respectively.

{
\newcommand{\fig}[1]{\frame{\includegraphics[width=0.11\hsize]{images/comparison_detection/#1}}}
\begin{figure*}[t]
    \centering
    \begin{tabular}{ccccccc}
        \fig{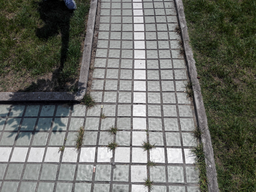} & \fig{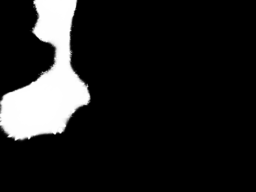} & \fig{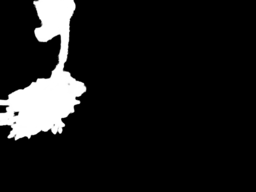} & \fig{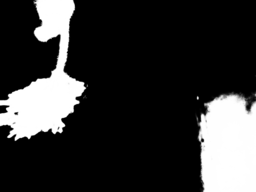} & \fig{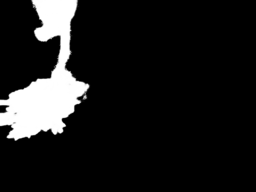} & \fig{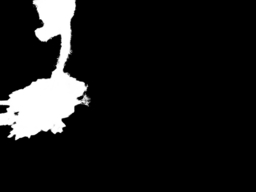} & \fig{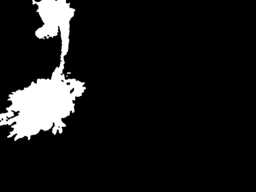} \\
        \fig{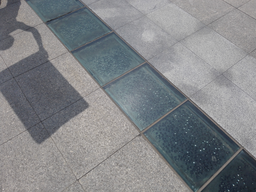} & \fig{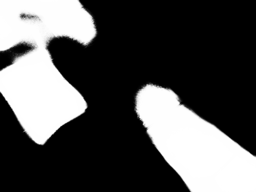} & \fig{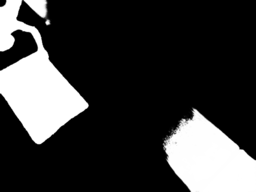} & \fig{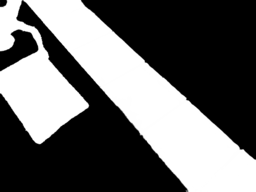} & \fig{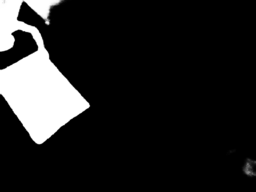} & \fig{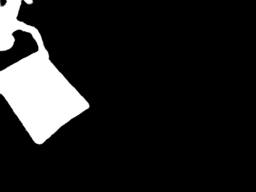} & \fig{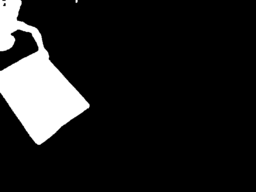} \\
        \fig{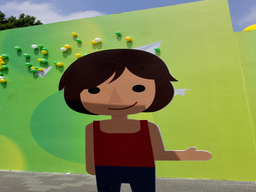} & \fig{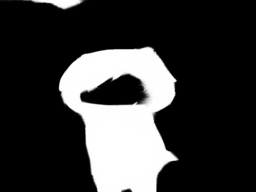} & \fig{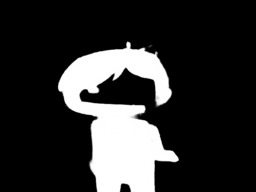} & \fig{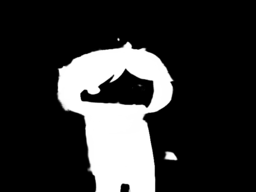} & \fig{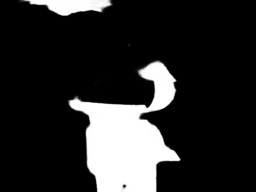} & \fig{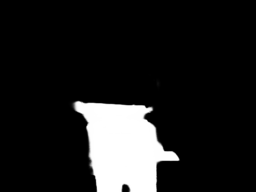} & \fig{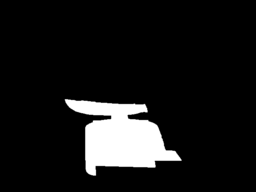} \\
        Input & DSC~\cite{hu2019direction} & BDRAR~\cite{zhu2018bidirectional} & DSDNet~\cite{zheng2019distraction} & BDRAR-ft & DSDNet++-ft & Ground truth \\
    \end{tabular}
    \caption{
        Comparison of shadow detection models evaluated on ISTD test set.
        BDRAR-ft and DSDNet++-ft denote BDRAR and DSDNet++ trained on SynShadow and fine-tuned on the ISTD train set, respectively.
    }
    \label{fig:comparison_detection}
\end{figure*}
}

\subsection{Experiments on ISTD dataset}
We first trained shadow detection models on SynShadow and then fine-tuned them on the ISTD dataset.
We show some example shadow detection results in \Fref{fig:comparison_detection}.
We observe that models trained on SynShadow and fine-tuned on the ISTD dataset can better distinguish correct shadow regions and challenging fake shadow regions such as tiles and dark areas by only employing pre-training and fine-tuning strategies.
The quantitative evaluation results are shown in \Tref{tbl:comparison_detection}.
The fine-tuned results clearly outperform the compared results, having almost 50$\%$ lower BER scores than the current best model, DSDNet.
Most of the improvement is attributed to non-shadow regions.

For more detailed analysis, we performed a comparison by changing the dataset for training and fine-tuning in \Tref{tbl:ablation_detection_istd}.
The model trained on SynShadow (2.74 BER in the ISTD dataset) is already comparable to most of the state-of-the-art models except the DSDNet model in \Tref{tbl:comparison_detection}.
We observe that SynShadow provides a better pre-trained model than GTAV~\cite{sidorov2019conditional}, SRD+, and SBU for fine-tuning on ISTD by comparing the fourth and fifth rows in each table.
This result supports our argument that SynShadow provides complementary information compared to the existing datasets.

We took the same procedure for the SBU dataset.
However, there was no performance improvement in BER.
We conjecture that this is due to the nature of the datasets.
Most of the shadows in the SBU dataset are caused by occluder objects visible in the camera view.
In contrast, many shadows in the ISTD dataset are caused by occluder objects invisible from the camera, which we also assume.

\begin{table}[t]
  \normalsize
  \setlength{\tabcolsep}{2pt}
  \caption{
      Quantitative shadow detection results evaluated on the ISTD testing set.
      Top two results in each setting are highlighted in \red{red} and \blue{blue}, respectively.
  }
  \label{tbl:comparison_detection}
  \centering
  \setlength{\tabcolsep}{2pt}
  \begin{tabular}{@{}cccc@{}} \toprule
      Method & S & NS & ALL \\ \midrule
      \multicolumn{4}{l}{\textit{Supervised}} \\
      Stacked-CNN~\cite{vicente2016large} & 7.96 & 9.23 & 8.60 \\
      scGAN~\cite{nguyen2017shadow} & 3.22 & 6.18 & 4.70 \\
      ST-CGAN~\cite{wang2018stacked} & 2.14 & 5.55 & 3.85 \\
      DSC~\cite{hu2019direction} & 3.85 & 3.00 & 3.42 \\
      BDRAR~\cite{zhu2018bidirectional} & \red{0.50} & 4.87 & 2.69 \\
      DSDNet~\cite{zheng2019distraction} & 1.36 & 2.98 & 2.17 \\ \midrule
      \multicolumn{4}{l}{\textit{Supervised, pre-trained on SynShadow}} \\
      BDRAR & \blue{0.62} & \blue{1.57} & \blue{1.10} \\
      DSDNet++ & 1.13 & \red{1.04} & \red{1.09} \\ \bottomrule
  \end{tabular}
\end{table}

\begin{table}[t]
  \normalsize
  \setlength{\tabcolsep}{2pt}
  \caption{
      Comparison by changing the dataset for training and fine-tuning of shadow detection models.
      Evaluation is performed on the ISTD testing set.
      Top two results in each setting are highlighted in \red{red} and \blue{blue}, respectively.
  }
  \label{tbl:ablation_detection_istd}
  \centering
      \begin{tabular}{@{}ccccccc@{}} \toprule
          Model & \multicolumn{3}{c}{DSDNet++} & \multicolumn{3}{c}{BDRAR} \\
          Metrics & S & NS & ALL & S & NS & ALL \\ \midrule
          \multicolumn{7}{l}{\textit{Training}} \\
          GTAV & 15.83 & 19.37 & 17.60 & 12.83 & 46.67 & 29.75 \\
          SRD+ & 10.68 & 2.36 & 6.52 & 4.63 & 8.64 & 6.64 \\
          SBU & 7.19 & \red{2.14} & 4.66 & 7.75 & \red{2.10} & 4.92 \\
          SynShadow & \blue{1.37} & 4.10 & \blue{2.74} & \blue{2.32} & \blue{3.17} & \blue{2.74} \\
          ISTD & \red{1.07} & \blue{3.01} & \red{2.04} & \red{0.50} & 4.87 & \red{2.69} \\ \midrule
          \multicolumn{7}{l}{\textit{Fine-tuning}} \\
          GTAV$\rightarrow$ISTD & 1.24 & \blue{1.98} & 1.61 & 0.74 & 2.89 & 1.81 \\
          SRD+$\rightarrow$ISTD & \red{0.74} & 2.89 & 1.81 & 0.41 & 3.85 & 2.13 \\
          SBU$\rightarrow$ISTD & \blue{1.02} & 2.07 & \blue{1.55} & \red{0.64} & \blue{2.55} & \blue{1.59} \\
          SynShadow$\rightarrow$ISTD & 1.13 & \red{1.04} & \red{1.09} & \blue{0.62} & \red{1.57} & \red{1.10} \\ \bottomrule
      \end{tabular}
\end{table}

\section{Discussion}
We discuss some of the limitations in SynShadow below.

\textbf{Objects inside the view}:
It cannot handle the shadow cast by objects inside the view.
However, we would like to argue that there are many applications where we want to remove shadows cast by objects outside the view, such as document and portrait shadow removal, as we discussed in \Sref{subsec:shadow_removal_methods}.

\textbf{Uneven surface}:
It assumes flat surfaces thus cannot explicitly cast shadows on uneven surfaces.
Viewers can sometimes find that the synthesized images look unnatural.
However, we can see some improved shadow removal results on uneven surfaces in \Fref{fig:comparison_removal_istd}.
Thus, shadow removal models can benefit from the proposed pipeline due to diverse shadows.

\textbf{Multiple lights}:
It is not straightforward to extend the shadow illumination model to work in environments where there are multiple primary lights, which can occur only in complex indoor scenes.
However, we believe that shadow removal is often required for avoiding strong shadow effects, which is usually caused by a single primary light like the sun or the brightest light in indoor scenes.

\section{Conclusion}
We presented SynShadow, a large-scale synthetic dataset of shadow/shadow-free/matte image triplets, by integrating the shadow illumination model, 3D models, and the shadow-free image collections.
This was enabled by introducing some assumptions such as occluder objects being outside the camera view and a flat surface for the shadow projection.
We showed that SynShadow is very useful.
SynShadow-trained shadow removal models outperformed existing approaches by achieving the best rating in a user study on the challenging USR dataset.
Fine-tuning SynShadow-pre-trained models achieved up to about $50\%$ BER reduction on the ISTD dataset in shadow detection, and up to about $10\%$ RMSE reductions on the ISTD+ and SRD+ datasets in shadow removal, compared to the best-performing approaches.
We hope our proposed shadow synthesis pipeline paves the way for future work to detect and remove diverse and challenging shadows, benefitting from the growing number of 3D models and shadow-free image collections.

\bibliographystyle{IEEEtran}

\begin{IEEEbiography}[{\includegraphics[width=1in,height=1.25in,clip,keepaspectratio]{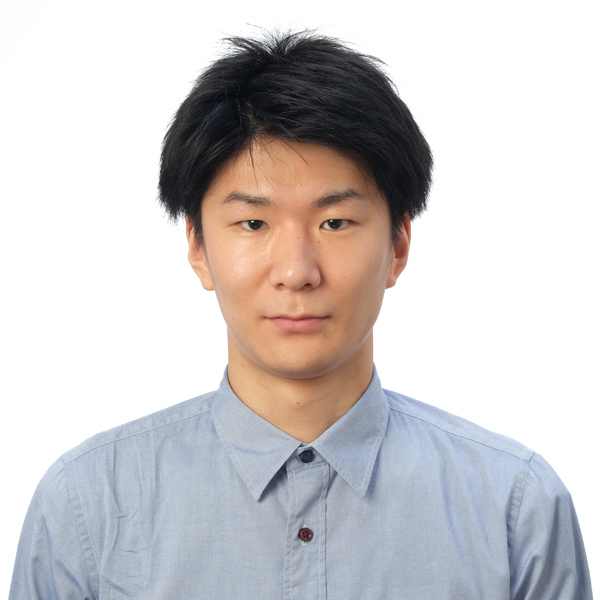}}]{Naoto Inoue}
  received the B.E. and M.S. in information and communication engineering from the University of Tokyo in 2016 and 2018, respectively. He is currently a Ph.D. student in The University of Tokyo, Japan. His research interests lie in computer vision, with particular interest in image generation and manipulation. He is a member of IEEE.
\end{IEEEbiography}
\vspace{-0.5cm}
\begin{IEEEbiography}[{\includegraphics[width=1in,height=1.25in,clip,keepaspectratio]{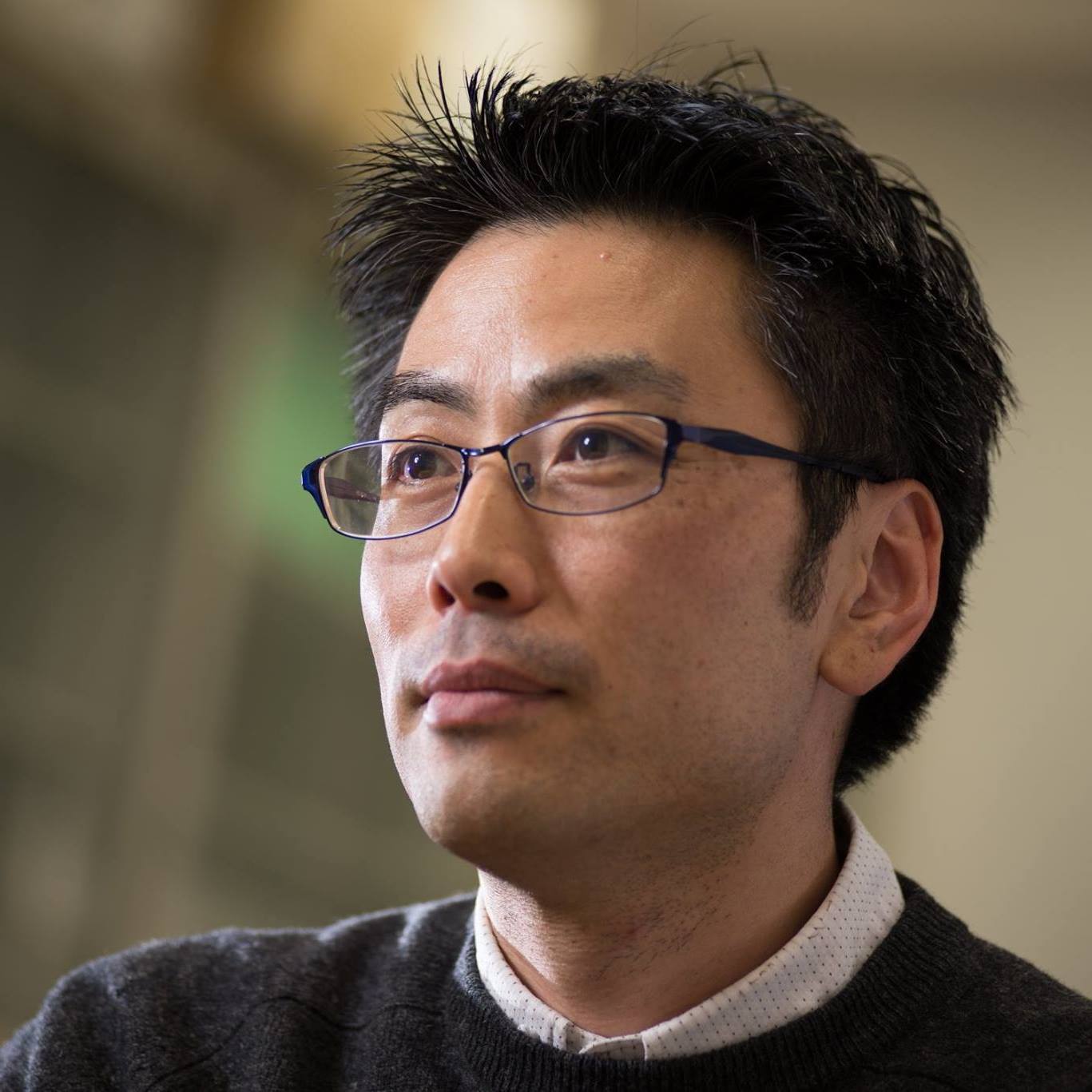}}]{Toshihiko Yamasaki}
received the B.S. degree in electronic engineering, the M.S. degree in information and communication engineering, and the Ph.D. degree from The University of Tokyo, Tokyo, Japan, in 1999, 2001, and 2004, respectively. He is currently an Associate Professor with the Department of Information and Communication Engineering, Graduate School of Information Science and Technology, The University of Tokyo. He was a Japan Society for the Promotion of Science Fellow for research abroad and a Visiting Scientist with Cornell University from 2011 to 2013. His has authored or coauthored three book chapters, more than 70 journal papers and more than 250 international conference papers. His research interests include multimedia big data analysis, pattern recognition, machine learning, etc. Prof. Yamasaki has received around 100 awards. He is a member of ACM, AAAI, IEICE, ITE, and IPSJ.
\end{IEEEbiography}

\end{document}